**Title**: HaptiNet: Networked Haptic Robots Enable Physical Co-presence in Geographically-Unconstrained Rehabilitation

**Authors:**

Chenyang Sun[1,2]†, Mingjie Dong[3]†, Haodong Deng[1], Yudong Liu[1], Yi-Feng Chen[1], Jun Lin[1], Changlong Huang[1], Jie Guo[1], Yantong Liu[1], Yang Liu[3], Yuzhou Lin[1], Jianjun Long[4], Zheng Xing[5], Sining Zhao[6], Xuemin Zhang[7], Zhiyong Wang[2], Zhenhong Li[8], Dongrui Wu[9], Honghai Liu[10], Jian S. Dai[11,12], and Mingming Zhang[1*]

**Affiliations:**

[1]Department of Biomedical Engineering, Southern University of Science and Technology, Shenzhen, 518055, China.

[2]School of Biomedical Engineering, Harbin Institute of Technology, Shenzhen, 518055, China.

[3]Beijing Key Laboratory of Advanced-Manufacturing Technology, College of Mechanical-&Energy Engineering, Beijing University of Technology, Beijing, 100124, China.

[4]Department of Rehabilitation, Shenzhen University of Advanced Technology General Hospital, Shenzhen, 518055, China.

[5]Department of Rehabilitation, Tianjin University Tianjin Hospital, Tianjin, 300211, China.

[6]School of Sports and Health, Tianjin University of Sport/Tianjin Key Laboratory of Sports Physiology and Sports Medicine, Tianjin 300381, China

[7]Rehabilitation Hospital, National Research Center for Rehabilitation Technical Aids, Beijing, 100124, China.

[8]School of Engineering, University of Manchester, Manchester M13 9PL, U.K.

[9]The Ministry of Education Key Laboratory of Image Processing and Intelligent Control, School of Artificial Intelligence and Automation, Huazhong University of Science and Technology, Wuhan, 430074, China.

[10]The University of Portsmouth, Portsmouth, PO1 3HE, UK.

[11]Department of Mechanical and Energy Engineering, Shenzhen Key Laboratory of Intelligent Robotics and Flexible Manufacturing Systems, Southern University of Science and Technology, Shenzhen, 518055, China.

[12]Centre for Robotics Research, School of Natural and Mathematical Sciences, King's College London, London, WC2R 2LS, U.K.

†These authors contributed equally to this work.

*Corresponding author. Email: zhangmm@sustech.edu.cn.

**Abstract:** Cooperative rehabilitation enhances engagement, task performance, and social-motor interaction, yet it demands physical co-presence: users must transmit forces, coordinate movements, and infer intent through haptic contact. Telerehabilitation promises to expand access for patients constrained by distance, mobility, or clinical disparities, yet current techniques remain predominantly audiovisual while leaving users haptically and physically isolated. Here, we introduce HaptiNet, a networked haptic robotic system enabling physical co-presence for geographically distributed users via force-mediated interaction. Each robotic terminal features a low-inertia, long-stroke design with high force-feedback capacity, tailored for haptic rendering in upper-limb training. Building on these terminals, HaptiNet creates a distributed haptic network with an imitation-learning-based delay compensator, enabling users to physically perceive and coordinate with one another over distance. We validated HaptiNet in 284 healthy participants and 111 patients with neurological impairments across progressively realistic settings, including laboratory tests, cross-city deployments, and clinical applications. HaptiNet preserved task-level force rendering consistency across single-user and multi-user scenarios. Compared with solo and visual cooperative training, haptic cooperation improved task performance by 24% and 22%, respectively, while also boosting engagement and interpersonal motor synchrony. Across three intercity links totaling approximately 4,000 km, HaptiNet maintained stable haptic interaction among patients with neurological impairments, producing a 3.87-fold greater baseline-to-training score improvement and a 106% higher patient-applied effort over the solo condition. These results establish HaptiNet as a promising route for multi-user cooperative training, extending haptically and socially connected rehabilitation care beyond co-located settings into geographically-unconstrained scenarios.

**One-Sentence Summary:** Networked haptic robots enable physical co-presence and improved clinical outcomes in geographically-unconstrained rehabilitation.

**Main Text:**

## INTRODUCTION

Neurological impairment extends beyond motor disability and is often accompanied by persistent social isolation (*1*). Deficits in motor and cognitive functions impede reintegration into everyday life and employment, reducing social participation and eroding interpersonal networks (*2, 3*). Social isolation is associated with adverse mental and physical health outcomes, including anxiety, depressive symptoms, cardiovascular risk, poor rehabilitation outcomes, and recurrent stroke (*4, 5*). Rehabilitation should therefore be viewed not only as functional restoration, but also as a process of rebuilding social connections (*6, 7*). However, existing rehabilitation paradigms remain predominantly individualistic, with limited emphasis on social interaction (*8-10*). Cooperative rehabilitation offers a promising alternative by providing shared training experiences, enhanced motor performance and social participation, and improving motivation and efficiency (*11, 12*).

Despite these advantages, cooperative rehabilitation has not yet been widely adopted as a routine training paradigm in clinical practice (*13-15*). Cooperative rehabilitation is most often implemented in person, where multiple patients train together in co-located settings and coordinate their movements through direct physical contact (*16-18*), shared object manipulation (*19-21*), or interconnected devices (*22-24*). However, in-person implementation remains constrained by geographical proximity, transportation access, and clinical resources. In contrast, remote implementation is more accessible, enabling social interaction across distance through game-based platforms or online meetings (*25-28*). However, current telerehabilitation systems confine patients to audiovisual interaction and provide limited support for force-mediated physical interaction among geographically distributed users.

This gap motivates robot-mediated haptic interaction, where users, though geographically separated, exchange forces through robots and thereby experience mutual haptic feedback (*29-30*). Such systems allow users to perceive partners' movement intentions via haptic connection, adjust their own actions accordingly, and coordinate effort within a shared sensorimotor loop (*31-33*). Previous studies have also shown that robot-mediated haptic interaction can improve joint task performance and promote motor learning in multi-user settings (*34-36*). However, existing haptic robotic systems remain inadequate for cooperative rehabilitation, as they have not yet jointly demonstrated the haptic-rendering capacity required for upper-limb training, the ability to preserve task-level force consistency for multi-user interaction, and the robustness for haptic transmission over distance.

Therefore, realizing haptic-mediated cooperative rehabilitation requires addressing two challenges. First, for each individual robotic terminal, haptic rendering must match the kinematics and dynamics of upper-limb motor tasks to support effective motor learning. Second, during cooperative tasks, haptic transmission must preserve sufficient fidelity to enable physical co-presence in the absence of co-location. Otherwise, distorted or delayed force feedback may undermine the benefits of cooperation. To address them, we developed HaptiNet, a networked haptic robot system for geographically-unconstrained cooperative rehabilitation. Each robotic terminal features a low-inertia, long-stroke design with high force-feedback capacity, enabling haptic rendering for upper-limb training. Building on this, HaptiNet interconnects multiple robotic terminals through a shared virtual environment to realize physical co-presence. Further, to preserve haptic fidelity despite network latency, we developed an imitation-learning-based motion predictor to compensate for latency-induced motion asynchrony. This architecture allows users to physically perceive one another's interaction forces and coordinate their movements within a shared training space, enabling haptic-mediated cooperative rehabilitation across geographic boundaries.

We systematically evaluated HaptiNet through a multi-stage, progressively realistic experimental framework involving 284 healthy participants and 111 patients with neurological impairments. The validation spanned robotic haptic-rendering fidelity, behavioral and physiological benefits of haptic cooperative rehabilitation, haptic transmission under network latencies, and cross-city clinical validation. Results demonstrate that HaptiNet maintained stable force transmission across single-user and multi-user conditions. Compared with solo training, haptic cooperation yielded clear behavioral and physiological evidence of cooperative benefits, reflected in improved task performance (24%), stronger engagement (subjective: 49%; objective: 27%), and enhanced interpersonal motor synchrony (16%). In the cross-city clinical validation spanning approximately 4,000 km, HaptiNet maintained stable haptic interaction and sustained cooperative benefits. The cooperative rehabilitation group showed a 3.87-fold greater baseline-to-training score improvement and a 106% greater patient-applied effort than the solo group. Our work extends the paradigm of cooperative rehabilitation from co-located to geographically-unconstrained, paving the way for next-generation networked haptic robotics that enables physical co-presence over distance.

## RESULTS

### HaptiNet preserves haptic-rendering fidelity

To assess the haptic rendering performance of HaptiNet, we examined whether the force required for a specific task remained consistent across different numbers of connected users (Fig. 1A; see Supplementary Methods and Movies for experimental details). Representative individual force profiles showed that users adopted different coordination strategies, either assisting one another in the same direction or intermittently applying opposing forces (Fig. S5). Despite these user-dependent interaction patterns, the resultant force in the two-user and three-user conditions remained consistent with the force generated in the single-user condition.

Group-level results across all ten triads are summarized in Fig. 1B for the three resistance levels. The gray, green, and red traces show the time-resolved mean force over the 20-s trial across the ten triads for the single-user, two-user, and three-user conditions, respectively. Shaded regions indicate ±1 standard deviation (SD). For the single-user condition, the force trace corresponds to the human–robot interaction force generated by the participant. For the multi-user conditions, the force trace corresponds to the resultant force generated by the participants. Across ten triads and three resistance levels, HaptiNet preserved the task-required force regardless of user number (Fig. 1B). Relative to the target resistance, the trial-averaged absolute force errors remained within 4% under the 4 N condition, within 2% under the 10 N condition, and within 3% under the 20 N condition across the single-user, two-user, and three-user conditions. The corresponding raw force values are provided in Table S1. These results demonstrate that HaptiNet maintained stable task-level force consistency across single-user and multi-user conditions, supporting its haptic-rendering fidelity for cooperative interaction.

### HaptiNet-mediated haptic coordination outperforms solo and vision-only conditions

We next examined whether haptic interaction could improve cooperative task execution beyond solo operation and visual-only coordination. Participants performed a tracking-and-catching task under three conditions: solo operation (Solo), visual-only multi-user coordination without force coupling (Visual co-op), and HaptiNet-mediated haptic coordination with force coupling (Haptic co-op) (Fig. 1C). This design allowed us to distinguish benefits arising from haptic interaction from those attributable to multi-user participation or shared visual feedback alone. Detailed experimental design is provided in Supplementary Methods and Fig. S6, and metric definitions are provided in Table S2-S3. We evaluated performance and effort using three objective measures (Fig. 1D) and assessed user experience using four subjective ratings (Fig. 1E).

Haptic co-op significantly improved task performance relative to the other two conditions. Compared with Solo and Visual co-op, Haptic co-op yielded significantly higher game scores ($P = 0.001$ and 0.005, respectively) and lower tracking errors ($P < 0.001$ for both), whereas Solo and Visual co-op did not differ significantly. Physical effort, quantified as the normalized applied force, was significantly greater in both multi-user conditions than in Solo ($P$ < 0.001 for both), indicating that multi-user participation itself promoted greater motor involvement. However, self-reported engagement was significantly enhanced only in Haptic co-op, suggesting that the additional subjective engagement was specifically associated with haptic interaction rather than greater physical effort alone.

Subjective questionnaire results showed that haptic cooperation increased engagement compared with both solo training and visual cooperation. Engagement was highest under haptic cooperation (4.59 ± 0.52), compared with solo training (3.63 ± 0.57, $P$ = 0.002) and visual cooperation (3.71 ± 0.65, $P$ < 0.001). Haptic realism was also higher under haptic cooperation than under visual cooperation (4.46 ± 0.49 vs. 3.73 ± 0.81, $P$ = 0.002), while remaining comparable to solo training (4.44 ± 0.42, $P$ = 1.000). Among the cooperation-specific ratings, perceived cooperation difficulty did not differ significantly between visual and haptic cooperation (4.24 ± 0.75 vs. 3.99 ± 0.99, $P$ = 0.472). In contrast, cooperation realism was significantly higher under haptic cooperation than under visual cooperation (4.70 ± 0.44 vs. 3.86 ± 0.91, $P$ < 0.001). The full statistic results are provided in Table S4-S7. The subjective results further indicate that performance improvement caused by haptic cooperation was also accompanied by greater engagement and a more realistic cooperative experience.

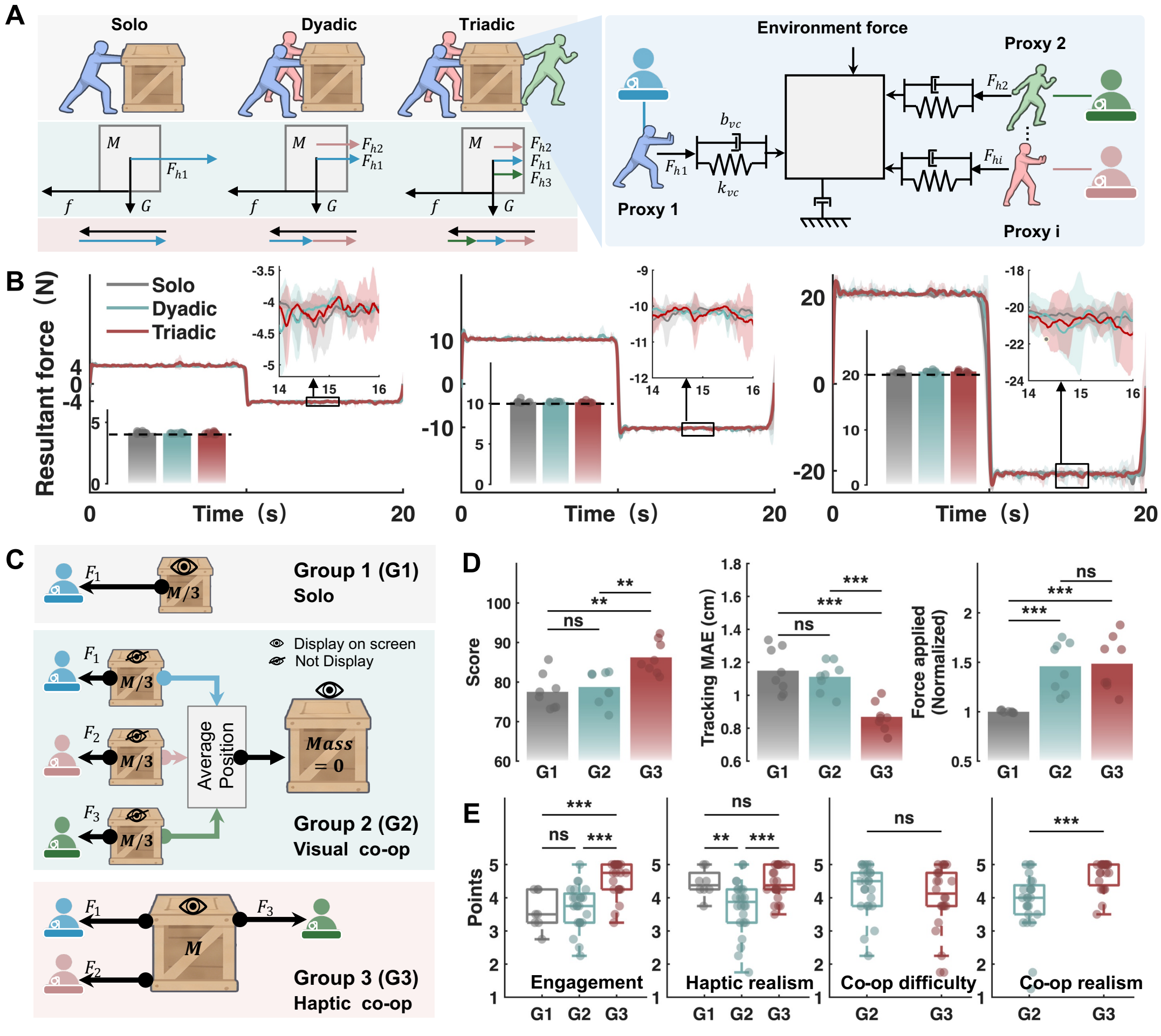

**Fig. 1. System performance (A-B) and behavioral validation of cooperative benefits (C-E).** (**A**) Experimental design for validating haptic transmission across different numbers of connected users. The left schematics illustrate the expected force equivalence among solo, dyadic, and triadic interaction, and the right schematic shows the HaptiNet-based haptic interaction architecture. $F_{hi}$: applied force from the $i$-th user; $k_{vc}$, $b_{vc}$: paremeters of virtual coupling. (**B**) Group-level resultant force profiles under three resistance levels. Solid lines indicate the mean, and shaded regions indicate ± 1 standard deviation (SD). For each resistance level, the upper-right inset shows a zoomed view of the force profile, and the lower-left inset shows the trial-averaged resultant force. (**C**) Experimental grouping for behavioral validation of cooperative benefits. The three groups were denoted as G1 to G3, corresponding to Solo, visual-only multi-user coordination without force coupling (Visual co-op), and HaptiNet-mediated haptic coordination with force coupling (Haptic co-op), respectively. (**D**) Objective measures of task performance and physical effort, including game score, tracking mean absolute error (MAE), and normalized applied force. Dots indicate individual participants for G1 and triads for G2–G3. (**E**) Subjective ratings of user experience, including engagement, haptic realism, perceived cooperation difficulty, and cooperation realism. Box plots show the median and interquartile range, with individual data points overlaid. ns, $P \geq 0.05$; *, $P < 0.05$; **, $P < 0.01$; ***, $P < 0.001$.

## Haptic interaction enhances physiological engagement and interpersonal synchrony

To determine whether HaptiNet-mediated haptic interaction modulates physiological states during cooperative training, we simultaneously recorded electroencephalography (EEG) and surface electromyography (sEMG) while participants performed the task under Solo and Haptic cooperative conditions (in the following sections, Co-op is used to refer to haptic cooperation unless explicitly noted). The experimental setup and protocol are shown in Fig. 2A and Fig. S7. Physiological engagement was quantified using the frontal engagement index (EI), defined as the β/α power ratio (13-30 Hz/8-13 Hz) at bilateral prefrontal electrodes F3 and F4. As bilateral prefrontal electrodes commonly used to index executive-control and sustained-attention demands, F3/F4 provide a task-related index of cognitive involvement. A higher β/α power ratio has been associated with greater prefrontal activation and attentional engagement during task performance (*37, 38*).

Compared with Solo, Co-op elicited significantly higher EI during the task period (Fig. 2B). Across the 3-13 s task interval, normalized EI increased at F3 (from 0.93 to 1.06; $P = 0.004$), F4 (from 0.94 to 1.06; $P = 0.037$), and the bilateral F3+F4 average (from 0.94 to 1.06; $P = 0.009$; Table S5). Difference topographies (Co-op minus Solo) further showed that this enhancement was concentrated over bilateral frontal regions (Fig. 2C), indicating that HaptiNet-mediated haptic interaction was associated with greater frontal physiological engagement than isolated solo performance. These EEG results suggest that the increased subjective engagement observed in the behavioral experiment was accompanied by a corresponding physiological signature.

We next examined whether haptic interaction influenced interpersonal motor coordination using sEMG recordings from eight upper-limb muscles (Fig. 2A). Compared with Solo, Co-op increased interparticipant sEMG coherence across multiple upper-limb muscles (Fig. 2D), indicating enhanced interpersonal motor synchrony during haptic interaction. As shown in Fig. 2E, group-level analyses in the 1-3 Hz band showed consistently higher coherence during Co-op in six of the eight muscles analyzed. These increases reached significance in biceps brachii (BB), anterior deltoid (AD), middle deltoid (MD), posterior deltoid (PD), middle trapezius (MT), and latissimus dorsi (LD), each showing the maximum signed-rank statistic given the six paired groups ($W = 21$, $P = 0.031$). The triceps brachii (TB) and pectoralis major (PM) showed no significant between-condition differences. For TB, this may be related to continuous grip-related activation while participants held the robot handle, which could reduce sensitivity to condition-specific coherence changes. Besides, the PM is anatomically close to the chest, and the normal human heart rate is approximately 60–100 beats per minute, corresponding to about 1.0–1.7 Hz. Because the statistical analysis in Fig. 2E focused on the 1–3 Hz frequency range, this band partially overlapped with the cardiac activity frequency range. Cardiac-related noise may therefore have affected the PM signal and partly explain the absence of a significant difference in the overall 1–3 Hz analysis. Notably, as shown in Fig. 2D, a significant difference in PM can be observed in the 2–3 Hz range. After excluding TB and PM, Co-op increased interparticipant sEMG coherence by an average of 16% compared with Solo. These results indicate that HaptiNet-mediated haptic interaction promoted coordinated muscle activity between users.

Together, the EEG and sEMG findings demonstrate that HaptiNet-mediated haptic interaction enhanced both physiological engagement and interpersonal motor synchrony. This multimodal evidence supports the behavioral results by showing that haptic cooperation not only improved task performance and perceived engagement but also elicited measurable muscular coordination across interacting users. The full statistic results are provided in Table S9 and S10.

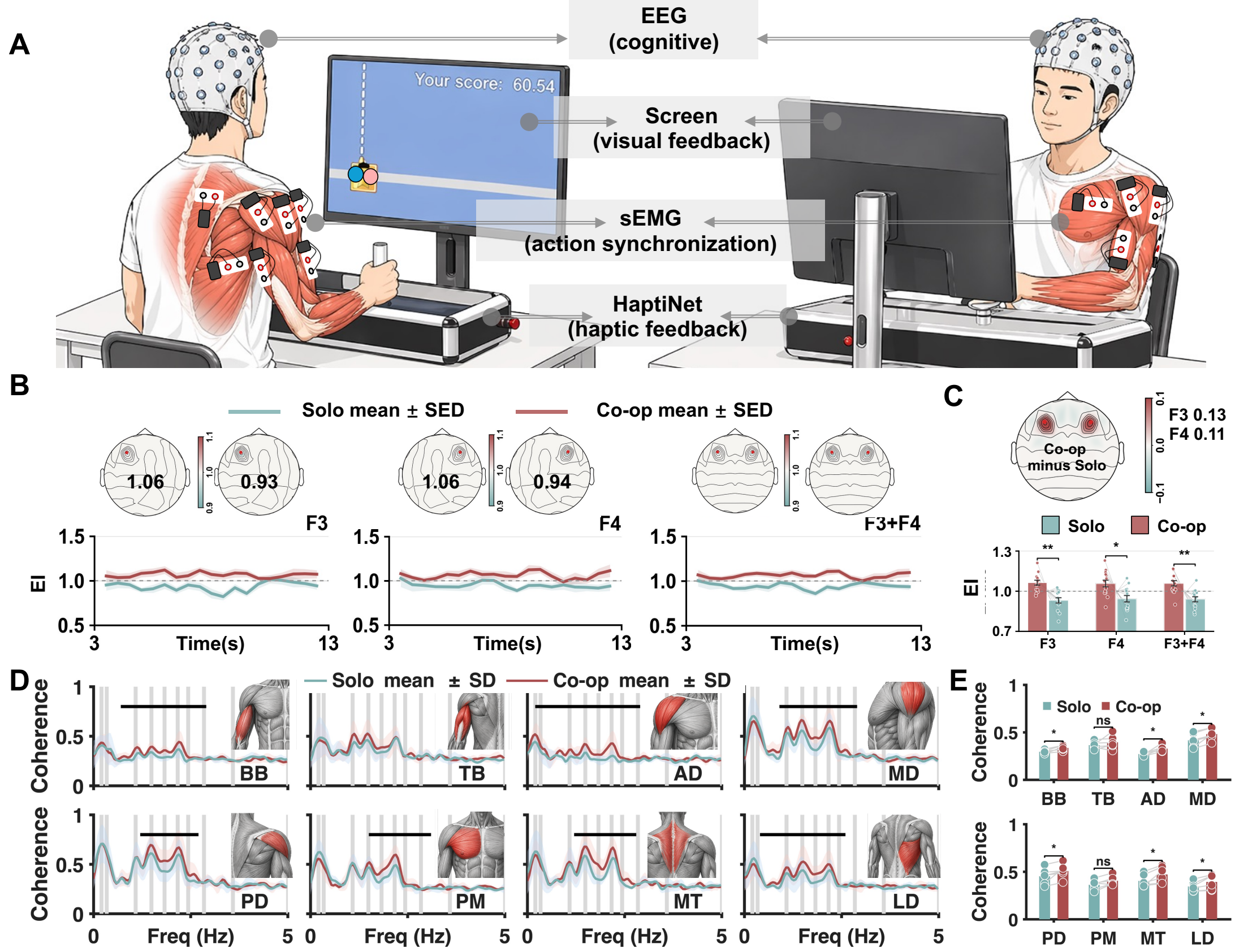

**Fig. 2. Physiological benefits of HaptiNet-mediated haptic cooperation.** (**A**) Experimental setup for assessing physiological responses. Electroencephalography (EEG) and surface electromyography (sEMG) were recorded to quantify physiological engagement and interpersonal action synchrony, respectively. (**B**) Time-resolved normalized engagement index (EI) at F3 channel, F4 channel, and their bilateral average. Topographic maps show the 3–13 s mean EI, with solo on the left and co-op on the right; red markers denote F3 and F4 channels. Lines indicate group means, and shaded regions indicate standard error of the mean (SEM) across participants. (**C**) Group-level 3–13 s mean EI. Bars show mean ± SEM, dots indicate individual participants, and the lower topography shows the co-op minus solo EI difference. (**D**) sEMG coherence of upper-limb muscles, showing interparticipant motor synchronization between paired users. Gray vertical bands indicate the frequency components of the tracked trajectory, and horizontal black bars denote frequency bands with significant differences between Solo and Co-op conditions. (**E**) Summary of mean sEMG coherence in selected muscles. Bars indicate mean ± SEM, and dots represent individual participants. Muscle abbreviations: BB, biceps brachii; TB, triceps brachii; AD, anterior deltoid; MD, middle deltoid; PD, posterior deltoid; PM, pectoralis major; MT, middle trapezius; LD, latissimus dorsi. ns, $P \geq 0.05$; *, $P < 0.05$, **, $P < 0.01$.

## Robust haptic interaction under network delays

As HaptiNet aims to enable geographically-unconstrained cooperative rehabilitation, network latency is an unavoidable factor that may affect remote haptic interaction. We therefore examined whether networked latency would degrade haptic transmission and cooperative benefits under controlled delay conditions. Participants performed the same tracking task under four latency conditions: 0 ms, 10 ms, 30 ms, and 50 ms, which were used to emulate transmission delays corresponding to local communication, local-area networks (LAN), metropolitan-area networks (MAN), and wide-area networks (WAN), respectively (Fig. 3A). The detailed experimental procedure is provided in Fig. S8.

The results in Fig. S9. showed that increasing latency affected both task score and resultant force. Specifically, task score decreased with increasing delay, indicating impaired cooperative task performance, whereas resultant force increased, suggesting that delayed haptic interaction required greater compensatory force to complete the same task. These findings confirm that network latency can compromise remote haptic cooperation and motivate the need for delay-compensation strategies in HaptiNet.

To mitigate latency-induced degradation in haptic transmission and cooperative benefits, we integrated a motion prediction module into HaptiNet (Fig. 3B). The module consisted of a velocity predictor and a state rollout component, which estimated the delayed remote state before rendering local haptic feedback. We evaluated three compensation conditions under the same simulated network latencies used above: no motion predictor (NMP), constant-velocity-based motion predictor (CVMP, Predictor 1), and imitation-learning-based motion predictor (ILMP, Predictor 2) (Fig. 3B). Detailed methods are provided in MATERIALS AND METHODS.

A representative 50-ms latency trial showed that, without prediction, the proxy of the remote partner lagged behind the true remote position, producing a pronounced mismatch (Fig. 3C). CVMP (Predictor 1) reduced the latency-induced mismatch but still produced overshoot around reversal points. In contrast, ILMP (Predictor 2) further suppressed reversal-related overshoot and generated a trajectory that more closely followed the true remote motion. Quantitative position-error analysis confirmed that prediction reduced latency-induced mismatch (Fig. 3D). At 10 ms, ILMP already reduced position error relative to NMP ($P = 0.006$), whereas CVMP did not differ significantly from NMP. At 30 and 50 ms, both CVMP and ILMP significantly reduced position error relative to NMP, with no significant difference between the two prediction-based methods.

We next assessed whether prediction preserved haptic force transmission under delay. At 0 ms, the three conditions produced comparable resultant forces. With increasing latency, however, resultant force in the NMP and CVMP conditions progressively deviated from the no-latency baseline, whereas ILMP remained closer to baseline (Fig. 3E). At 10 ms, both CVMP and ILMP differed significantly from NMP, and at 30 and 50 ms, all pairwise comparisons were significant. Force-ratio analysis further showed that the latency-dependent increase in resultant force was strongest in NMP, reduced in CVMP, and nearly absent in ILMP, with slopes of 0.193, 0.051, and 0.009, respectively (Fig. 3F). These results indicate that ILMP most effectively stabilized haptic force transmission across latency conditions.

Finally, task-score analysis further demonstrated the advantage of ILMP (Fig. 3G). At 10 ms, the compensation strategy had no significant effect on the score. At 30 ms, ILMP achieved higher scores than NMP ($P$ = 0.005). At 50 ms, both CVMP and ILMP outperformed NMP, and ILMP further outperformed CVMP ($P$ = 0.011). Together, these results show that although CVMP and ILMP produced similar average position-error reductions at higher networked latencies, ILMP better suppressed reversal-related overshoot, preserved resultant force, and maintained task performance. Thus, compared with CVMP, ILMP more effectively mitigated latency-induced degradation in haptic transmission and cooperative task performance. Full results of statistical analysis are provided in Table S11-S14.

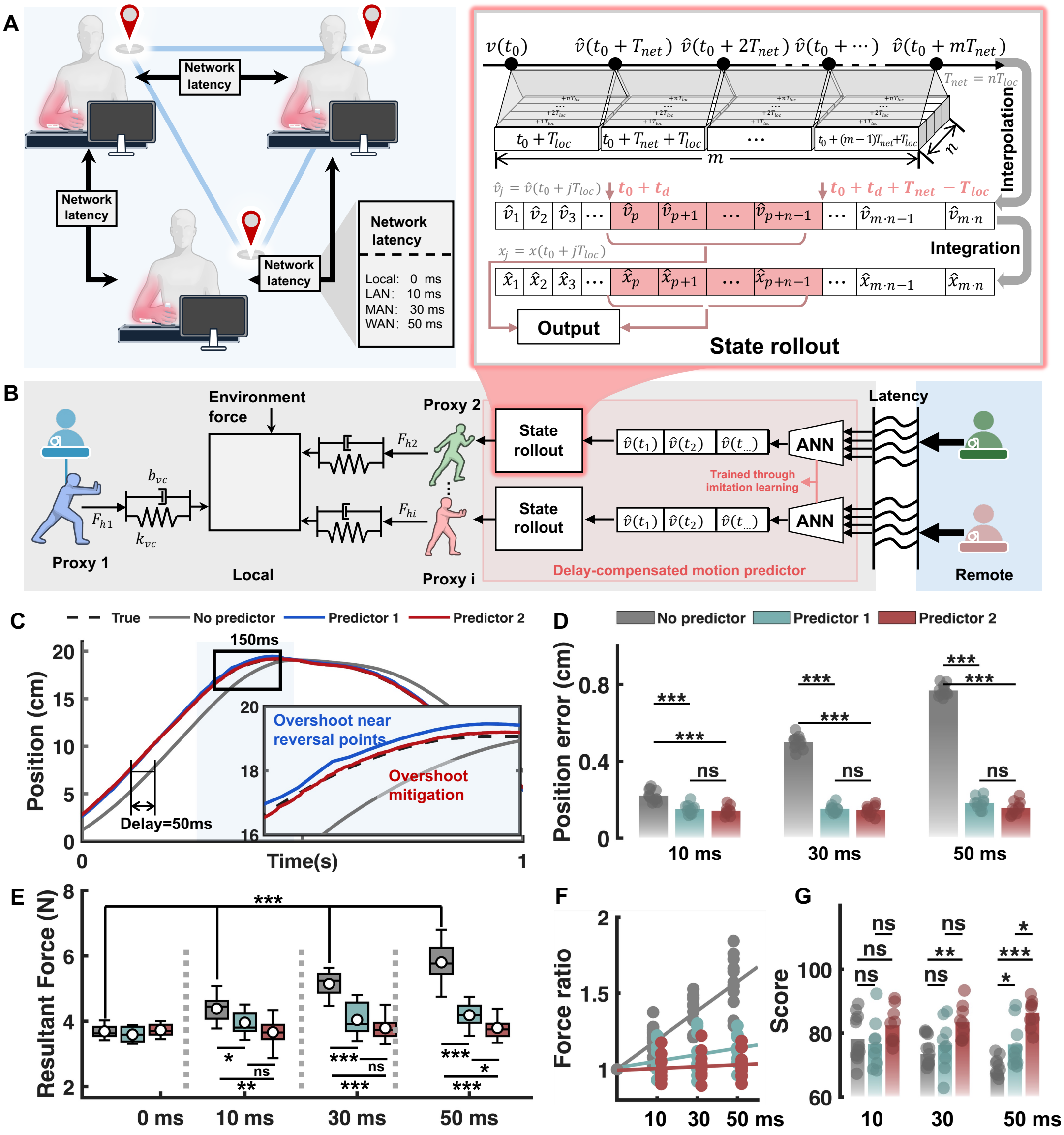

**Fig. 3. Robustness of HaptiNet-mediated haptic interaction under network latencies**. (**A**) Experimental network-latency configuration. Haptic interaction was evaluated under four emulated latency conditions: local interaction without delay (0 ms), local-area network (LAN, 10 ms), metropolitan area network (MAN, 30 ms), and wide-area network (WAN, 50 ms). (**B**) Control framework for delay-compensated haptic interaction. The predictor uses an artificial neural network (ANN) trained through imitation learning to predict the remote user's velocity online, followed by state rollout to estimate the current remote motion state from delayed

information. $F_{hi}$: applied force from the $i$-th user; $k_{vc}$, $b_{vc}$: paremeters of virtual coupling; $T_{net}$ : network update period; $T_{loc}$ : local update period; $m$ : steps of ANN output; $n = T_{net}/T_{loc}$ , ($n \in Z^{+}$); $t_d$ : measured network latency. (**C**) A representative tracking trajectory under a 50-ms delay, showing the remote true trajectory and the locally rendered trajectory under three compensation conditions: No predictor, constant-velocity-based motion predictor (CVMP, Predictor 1), and imitation-learning-based motion predictor (ILMP, Predictor 2). (**D**) Position tracking error under different network latencies and compensation conditions. Bars indicate means, and dots indicate individual triads. (**E**) Resultant force under different network latencies and compensation conditions. Box plots show the median and interquartile range. (**F**) Force ratio across network-delay conditions, defined as the resultant force at 10 - 50 ms divided by the resultant force at 0 ms within the same compensation condition. (**G**) Task score under different conditions. Bars indicate means, and dots indicate individual triads. ns, $P \geq 0.05$; *, $P < 0.05$; **, $P < 0.01$; ***, $P < 0.001$.

### Long-distance deployment of HaptiNet across 4000 km

To further validate HaptiNet under remote conditions, we deployed the system across three geographically separated sites and connected participants through HaptiNet (Fig. 4A). The three intercity links showed distinct latency distributions, with delays of approximately 34.5, 40.1, and 43.8 ms for the A–B, A–C, and B–C connections, respectively (Fig. 4B). Participants completed baseline solo testing, remote cooperative training, and post-training testing (Fig. 4C). During the remote sessions, they performed the tracking task under three randomized conditions: Solo, Co-op 1 using CVMP, and Co-op 2 using the ILMP.

Task scores were analyzed during both training and post-training test phases. Co-op 2 achieved the highest task scores in both phases (Fig. 4D). Repeated-measures ANOVA showed a significant main effect of condition during both training ($P < 0.001$) and post-training test phases ($P < 0.001$). Post hoc comparisons showed that Co-op 2 significantly outperformed both Solo and Co-op 1 during practice ($P < 0.001$ for both) and test phases ($P < 0.001$ for both). These results indicate that ILMP better preserved remote cooperative task performance than CVMP under real intercity network latency. Tracking MAE showed a consistent pattern (Fig. S10).

Subjective questionnaire results further supported the advantage of ILMP during long-distance haptic cooperation (Fig. 4E). For engagement, statistical analysis showed that Co-op 2 produced significantly higher engagement than Solo ($P = 0.003$) and Co-op 1 ($P < 0.001$), whereas Solo and Co-op 1 did not differ significantly ($P = 0.445$). For haptic realism, Co-op 1 showed significantly lower haptic realism than Solo ($P = 0.003$) and Co-op 2 ($P = 0.012$), whereas Co-op 2 did not differ significantly from Solo ($P = 0.539$). This pattern suggests that ILMP helped preserve the perceived realism of haptic feedback under intercity network latency. Because Co-op difficulty and Co-op realism were specific to cooperative interaction, these two items were compared only between Co-op 1 and Co-op 2. Co-op 2 showed significantly better ratings for the Co-op difficulty item ($P = 0.047$) and significantly higher Co-op realism ($P < 0.001$) than Co-op 1. Together, these questionnaire results indicate that ILMP improved not only task performance but also the subjective quality and realism of remote haptic-mediated cooperation.

Force analysis further showed that ILMP better preserved haptic-rendering fidelity during long-distance cooperation (Fig. 4F). For the resultant force $|F_{res}|$, repeated-measures ANOVA revealed a significant effect of condition in both phases ($P < 0.001$). Post hoc comparisons showed that the ILMP condition did not differ significantly from the Solo baseline, whereas the CVMP condition differed significantly from both Solo and ILMP ($P < 0.001$). In contrast, the average absolute interaction force $|F_h|$ of each participant did not differ significantly between CVMP and ILMP. Full results of statistical analysis are provided in Table S15-S17.

The radar summary further shows that ILMP outperformed the other conditions across multiple dimensions, including task performance, haptic-rendering fidelity, and subjective experience (Fig. 4G). Together, these results demonstrate that HaptiNet can support remote haptic-mediated cooperation across intercity distances. More importantly, ILMP enabled more stable and realistic remote cooperation than CVMP, supporting the feasibility of HaptiNet for geographically-unconstrained cooperative rehabilitation.

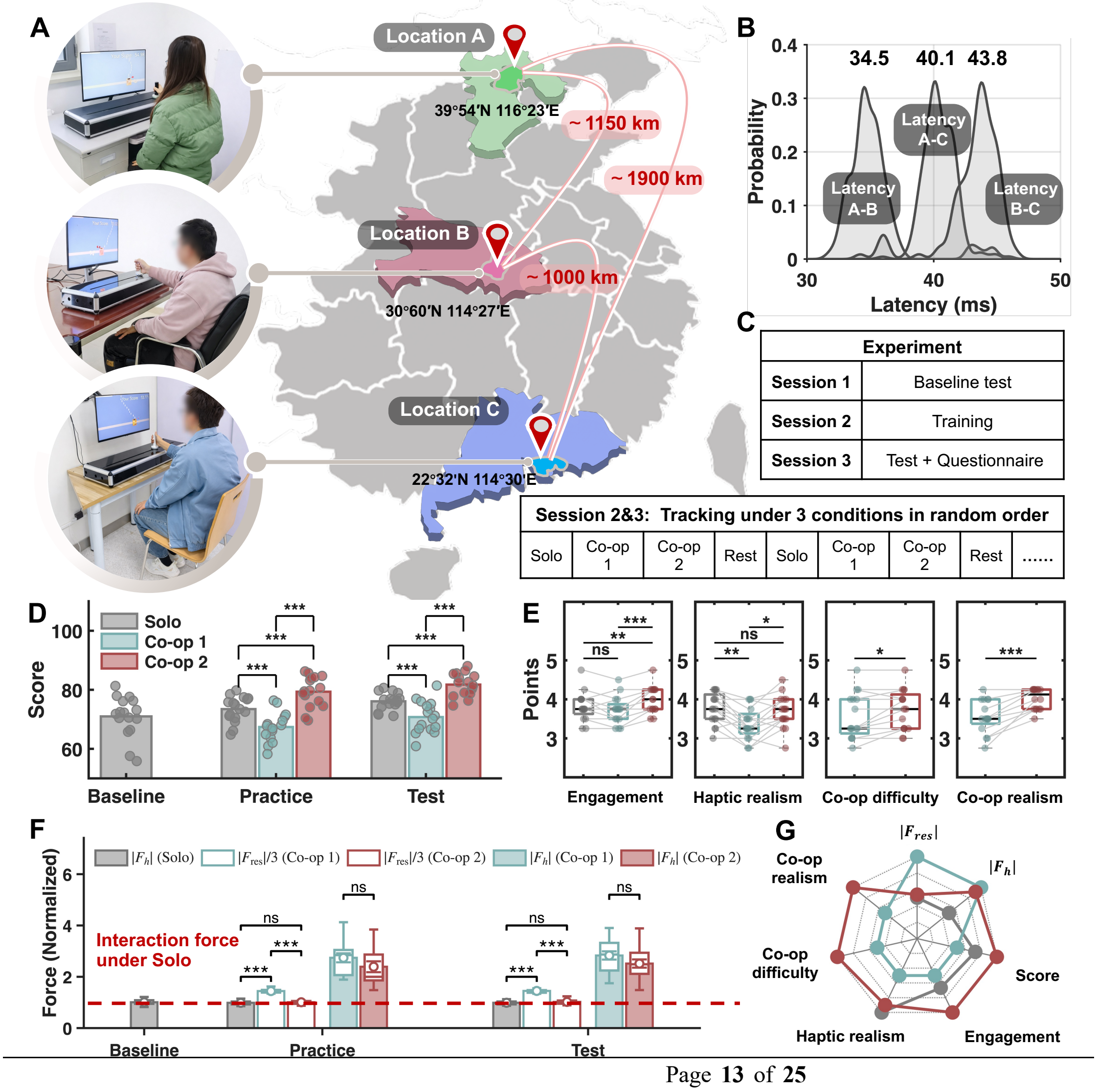

**Fig.4. Remote validation of HaptiNet for long-distance haptic cooperative rehabilitation.** (**A**) Deployment of HaptiNet across three geographically separated sites, denoted as Locations A-C, with the latitude and longitude of each city annotated on the map. (**B**) Measured network latency distributions among the three remote links, showing distinct delay profiles for A–B, A–C, and B–C connections. (**C**) Experimental protocol. Participants completed baseline solo testing, remote cooperative training, and post-training testing with questionnaires. In Sessions 2 and 3, the tracking task was performed under randomized Solo, Co-op 1, and Co-op 2 conditions. Co-op 1 used CVMP, whereas Co-op 2 used ILMP for latency compensation. (**D**) Task performance scores across baseline, training, and post-training test phases. Bars indicate means, and dots indicate individual triads. (**E**) Subjective questionnaire results, including engagement, haptic realism, cooperation difficulty, and cooperation realism. Box plots show the median and interquartile range, with individual triads data points overlaid. Data belonging to the same individual triads are linked. (**F**) Force analysis. $|F_h|$ denotes the normalized average absolute interaction force of each participant, and $|F_{res}|$ denotes the normalized average resultant force generated by the group. Forces were normalized to $|F_h|$ during the baseline period. Box plots show the median and interquartile range. (**G**) Radar summary of task performance, interaction force, resultant force, and subjective experience. ns, not significant; *, $P < 0.05$; **, $P < 0.01$; ***, $P < 0.001$.

## Clinical validation in patients with neurological impairments

To evaluate the clinical feasibility of HaptiNet-mediated cooperative rehabilitation, we conducted patient experiments under both local and remote deployments (Fig. 5A and B). Patients were included in either the Solo training group or the HaptiNet-mediated cooperative rehabilitation group (Co-op group), which consisted of local and remote cooperative sessions (Fig. 5C). A total of 111 patient were included in this study, the majority of whom were post-stroke patients. Other neurological conditions included brain tumors, moyamoya disease, and related disorders. Detailed demographic and clinical characteristics of the patients are provided in Fig. S11.

Fig. 5D shows the scatter distributions of score versus applied force under the Solo and Co-op conditions during the baseline and training phases, with score on the x-axis and applied force on the y-axis. Fig. 5E presents the trial-by-trial results of score, tracking MAE, and applied force. Fig. 5F summarizes the phase-level averages. At baseline, task performance was comparable between Solo and Co-op. Specifically, baseline score was 55.65 in Solo and 55.43 in Co-op, with no significant difference ($p = 0.880$), while baseline tracking MAE was 2.59 cm and 2.80 cm, respectively, which was also not significantly different ($P = 0.467$).

As training progressed, task performance improved in both conditions, with larger improvements in the Co-op condition. For score, Solo increased from 55.65 during baseline to 58.97 during training, corresponding to an improvement of 3.32 ($p = 0.001$), whereas Co-op increased from 55.43 to 68.24, corresponding to an improvement of 12.81 ($P < 0.001$). For tracking MAE, Solo decreased from 2.59 cm to 2.27 cm, a reduction of 0.33 cm ($P < 0.001$), whereas Co-op decreased from 2.80 cm to 1.92cm, a reduction of 0.88 cm ($P < 0.001$). Together, these results indicate that although both groups showed improved task performance after training, the magnitude of improvement was larger in Co-op, as reflected by both the larger increase in score and the greater reduction in tracking MAE.

In this study, the applied force was used as an indicator of patient effort. During baseline, the force applied was comparable between Solo and Co-op (2.70 N and 2.58 N, respectively) and no significant difference was observed ($P$ = 0.134). In contrast, during training, the force applied was markedly higher in Co-op than in Solo, reaching 5.67 N and 2.76 N, respectively ($P$ < 0.001). This result suggests that the Co-op condition significantly enhanced patients' active engagement and motor output during training. The statistical results for tracking MAE are shown in Fig. S12, and the full results of statistical analysis are provided in Table S18.

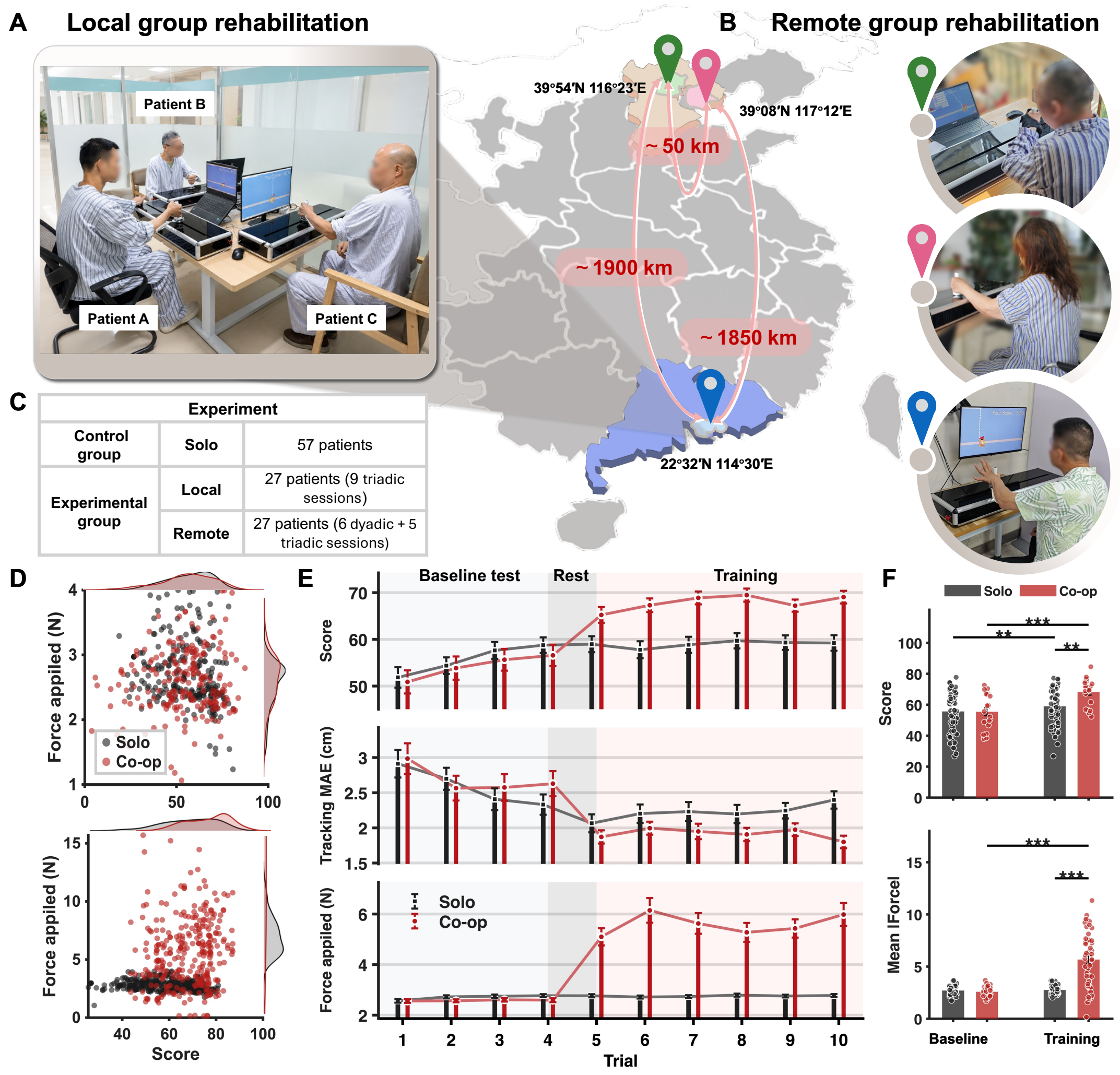

**Fig.5. Clinical validation of HaptiNet in patients with neurological impairments. (A)** Cooperative rehabilitation deployment at the same location, with three patients performing HaptiNet-mediated cooperation training simultaneously. **(B)** Cross-geographic validation setting, where the remote patient experiments were conducted across three sites, denoted as Locations A-C, with the latitude and longitude of each city annotated on the map. **(C)** Experimental grouping. Patients in the Solo group received solo rehabilitation training, whereas patients in the Co-op group received HaptiNet-mediated cooperative rehabilitation training, including local and remote Co-op sessions with dyadic or triadic haptic cooperation. (**D**) Distribution of task score and interaction force in Solo and Co-op conditions, with the upper panels showing the baseline period and the lower panels showing the training period. (**E**) Trial-by-trial changes in score, tracking MAE, and interaction force across baseline and training phases. (**F**) Comparison of task score and interaction force between the control and experimental groups. Bars indicate means. For Solo score, Solo mean force, and Co-op mean force, dots indicate individual participants. For Co-op score, dots indicate individual triads. All pairwise comparisons among the four conditions were performed. Only significant comparisons are annotated in the figure; unlabeled comparisons indicate no significant difference. *, $P < 0.05$; **, $P < 0.01$; ***, $P < 0.001$.

## DISCUSSION

Here, we present HaptiNet, a networked haptic robotic system for geographically-unconstrained cooperative rehabilitation. This system enables multiple geographically distributed patients to share interaction forces, perceive mutual intentions, and coordinate movements within a common haptic environment. The success of our approach hinges on two core components. The first is a distributed networked haptic architecture composed of multiple low-inertia, long-stroke robotic terminals, each providing sufficient force-feedback capacity for haptic rendering of upper-limb rehabilitation. The second is an imitation-learning-based motion prediction strategy that compensates for delayed motion states, reducing the detrimental effects of network latency on haptic fidelity and cooperative benefits. Together, these components enable stable and delay-compensated haptic cooperation among geographically distributed patients.

To validate the system, we conducted a series of experiments that progressed from basic system performance to realistic application scenarios. Overall, these experiments were organized into three stages. The first stage evaluated HaptiNet's fundamental capabilities in haptic rendering and transmission, as well as the benefits of haptic-mediated cooperation. We first assessed the accuracy and stability of force rendering across different numbers of users and force levels, demonstrating that the system maintained consistent haptic output in both single-user and multi-user conditions. Next, we validated the behavioral benefits of cooperative training. Participants executed a tracking-and-catching task under Solo, Visual co-op, and Haptic co-op conditions; the Visual co-op condition specifically served to rule out confounding factors such as visual sharing, task co-presence, and social companionship. The results showed that Visual co-op did not improve task performance, whereas haptic interaction led to higher task scores, lower tracking errors, and stronger perceived cooperation and engagement. We further compared Solo and Haptic Co-op using EEG and sEMG signals. EEG results indicated increased task engagement during Haptic Co-op, while sEMG results revealed enhanced coordinated motor activity between users. Collectively, these results show that HaptiNet establishes a high-fidelity haptic channel through which collaboration shifts from visual or verbal interaction to embodied physical co-presence.

The second stage examined the robustness of haptic-mediated cooperation under network delay. Remote haptic interaction is inevitably affected by communication latency, which can compromise the immediacy and stability of force feedback. Our experiments showed that, without delay compensation, increasing latency progressively distorted haptic rendering, reduced interpersonal coordination, and weakened the cooperative advantage, in some cases causing performance to fall below that of solo training. To address this problem, we developed an ILMP-based delay-compensation strategy that utilizes delayed information received from the remote terminal to estimate the current, real-time motion states of the remote partner. This strategy effectively mitigated delay-induced haptic distortion, preserved interaction stability, and maintained the benefits of haptic-mediated cooperation under delays up to 50 ms. Building on this latency-compensation framework, we further evaluated HaptiNet across three intercity links spanning approximately 4,000 km. Results from 48 healthy participants confirmed that HaptiNet could support stable and safe haptic interaction between geographically separated users.

Building on previous results, the third stage extended validation to application-oriented rehabilitation scenarios. We deployed the system in clinical trials involving patients with neurological impairments under both co-located and remote cooperative rehabilitation paradigms. The results from 111 patients were consistent with the trends observed in the preceding healthy-participant experiments, indicating that haptic cooperation produced greater training benefits. In terms of task score, the Solo condition improved by 6% from baseline to training, whereas the Co-op condition improved by 23%, corresponding to an approximately 3.87-fold greater improvement than Solo. The results further showed that haptic cooperation enhanced patients' active participation and motor output during training. The cooperation rehabilitation group showed a 106% greater patient-applied effort than the solo group. This is particularly important for rehabilitation, because patients' active participation and sustained effort often directly affect training quality and the potential benefits of rehabilitation.

It is worthwhile mentioning that the current results mainly reflect immediate improvements in task performance and patient-applied effort during training, and should not be directly equated with long-term motor functional recovery. Future studies with larger cohorts, longer intervention periods, and more rigorous randomized controlled designs are needed to evaluate the effects of HaptiNet-mediated cooperative rehabilitation on clinical scales. In addition, variations in disease type, impairment severity, and inter-patient interaction strategies may influence the efficacy of haptic cooperation, underscoring the need to identify which patient populations benefit most from this paradigm.

Overall, this study demonstrates that HaptiNet supports remote haptic-mediated cooperative rehabilitation, with enhanced task performance, interpersonal motor synchrony, and training engagement. These results support HaptiNet as a promising platform for multi-user cooperative training, extending haptically and socially connected rehabilitation care beyond co-located settings into geographically-unconstrained scenarios.

## MATERIALS AND METHODS

### System Development

In HaptiNet, each robotic terminal acts as an interaction medium, conveying a user's interaction forces and motion to other connected users. The system architecture integrates four key components.

i. Robotic device. To meet the combined requirements of low inertia, high force output, and lightweight design, the transmission mechanism and motion range were jointly optimized. Timing-belt transmission was selected over commonly used schemes (e.g., gears, chains, and ball screws) to facilitate low reflected inertia and a lightweight structure; compared with cable-driven mechanisms, timing belts further provide improved wear resistance, reduced risk of breakage, and higher load-carrying capacity (*39, 40*). The motion range was determined based on anthropometric considerations: adult shoulder-to-wrist length is approximately 0.332 times body height, yielding an arm length of ~600 mm for an adult of ~1.8 m height (*41*). Accordingly, the robotic device was designed with an effective travel range of 600 mm. The handle and force sensor form the end-effector interface, enabling direct physical interaction between the user and the robot. To accommodate patients with limited upper-limb strength, a forearm support was integrated at the end-effector. Detailed descriptions of the robotic devices are provided in the Supplementary Methods.

ii. Control framework for local haptic rendering. Three task modes were involved in this study: no-load task, single-user loaded task, and multi-user cooperative task. The no-load task corresponded to free movement when the user was not in contact with the virtual object in the virtual environment. The single-user loaded task and the multi-user cooperative task corresponded to the Solo condition and the haptic cooperation condition, respectively. Detailed control implementations for each task mode are provided in the Supplementary Methods.

iii. Shared virtual environment synchronization module. To support stable remote multi-user haptic interaction (haptic network), we further developed a distributed shared virtual environment and synchronization module based on the local haptic rendering framework described in ii. Each terminal within the haptic network maintained a local virtual environment, where the shared task object and remote partner proxies are replicated locally, enabling real-time interaction with the task scene and remote partners while preserving local responsiveness. Detailed descriptions are provided in the following sections.

iv. ILMP-based delay compensator. This module was designed to compensate for network-induced delays during remote haptic interaction. By predicting the current states of remote partners from delayed motion information, the delay compensator reduces latency-induced motion asynchrony. Detailed implementation is provided in the following sections.

### Shared virtual environment synchronization

HaptiNet adopted a distributed local virtual environment architecture to support remote multi-user haptic cooperation. The system consisted of $n$ user terminals. Each terminal was equipped with a local robotic device and a local virtual environment. The local virtual environment at the $i$-th terminal was denoted as $E_i$, where $i = 1,2,\cdots,n$. Each local virtual environment maintained local representations of all $n$ users' virtual proxies, together with a local dynamic replica of the shared task object. The virtual proxy of the $j$-th user represented in $E_i$ was denoted as $P_{j,i}$. When $j = i$, $P_{i,i}$ represented the local virtual proxy directly controlled by the robotic handle at the $i$-th terminal. When $j \neq i$, $P_{j,i}$ represented the synchronized copy of the $j$-th remote user's virtual proxy in $E_i$. In this way, each terminal maintained a complete multi-user virtual task scene, allowing the local user to interact with both remote users' virtual proxies and the shared task object.

The local virtual proxy position of the $i$-th user was denoted as $x_{p,i,i}$. The state of the local virtual proxy was transmitted to the other terminals through the network. For the $i$-th terminal, the synchronized copy of the $j$-th user's virtual proxy was represented as $x_{p,j,i}(t) = \hat{x}_{p,j\to i}(t)$, where $\hat{x}_{p,j\to i}(t)$ denote the virtual proxy state transmitted from terminal $j$ to terminal $i$.. Due to network transmission, the received state may be delayed with respect to the current remote state, and this delay can be compensated for by motion prediction.

The shared task object was modeled as a virtual box with a mass $M$. The local object replica in $E_i$ was denoted as $O_i$, with position given by $x_{o,i}$. In each local virtual environment, all users' virtual proxies were connected to the local object replica through virtual coupling. The virtual coupling was implemented as a spring-damper connection that converted the relative motion between a virtual proxy and the object into an interaction force applied to the object. In $E_i$, the virtual-coupling force generated by $P_{j,i}$ on the local object replica $O_i$ was defined as

$$F_{j\to o,i}^{vc} = k_{vc}\left(x_{p,j,i} - x_{o,i}\right) + b_{vc}\left(\dot{x}_{p,j,i} - \dot{x}_{o,i}\right) \tag{1}$$

where $k_{vc}$ and $b_{vc}$ are the stiffness and damping coefficients of the virtual coupling between the user proxy and the local object replica, respectively.

The task-level interaction force applied to the local object replica in $E_i$ was computed as the sum of the virtual-coupling forces generated by all users:

$$F_i^{task} = \sum_{j=1}^{n} F_{j\to o,i}^{vc} \tag{2}$$

where $F_i^{task}$ is the task-level force generated by all users through virtual coupling. Because each terminal independently maintained a local object replica, network delay, packet loss, prediction error, and numerical integration error could cause state inconsistency among replicas. To maintain consistency of the shared object across distributed virtual environments, a synchronization coupling was introduced among object replicas. The synchronization correction force in $E_i$ was defined as

$$F_i^{sync} = \frac{1}{n-1}\sum_{j\neq i}[k_{sync}(\hat{x}_{o,j\to i} - x_{o,i}) + b_{sync}(\hat{\dot{x}}_{o,j\to i} - \dot{x}_{o,i})] \quad (3)$$

where $F_i^{sync}$ is the synchronization correction force among object replicas, $\hat{x}_{o,j\to i}$ and $\hat{\dot{x}}_{o,j\to i}$ are the position and velocity states of the object replica transmitted from terminal $j$ to terminal $i$, respectively, $k_{sync}$ and $b_{sync}$ are the stiffness and damping coefficients of the synchronization coupling among object replicas. The factor $1/(n-1)$ was used to normalize the synchronization correction with respect to the number of remote replicas. Similar to $\hat{x}_{p,j\to i}(t)$, the received object state may be delayed and can be compensated by motion prediction.

The dynamics of the local object replica in $E_i$ were given by

$$M\ddot{x}_{o,i} = F_i^{task} + F_i^{sync} + F_i^{env} \quad (4)$$

where $F_i^{env}$ represents other task-related forces in the virtual environment, such as friction, resistance, virtual boundary constraints, or external loads. It should be noted that the synchronization correction force was used to reduce state deviations among object replicas and was not interpreted as a task-level interaction force generated by the users. Thus, the object motion was primarily determined by the combined virtual-coupling forces generated by all users, while the synchronization term maintained consistency of the shared object state across terminals.

**Model training for ILMP**

To predict future system states over a short horizon for latency compensation, we developed an ILMP-based delay compensator. The ILMP was designed to learn task-related motion patterns from recorded user–object interactions and to estimate future velocity trajectories from recent motion histories.

During model training, motion data collected from single-user loaded movement tasks were used as input–output samples, allowing the predictor to capture the relationship between user motion, virtual object motion, and target trajectory information. In HaptiNet, the local control loop updates at 1000 Hz, corresponding to a local control period of $T_{loc}$=1 ms. The networked state update operated at 100 Hz, corresponding to a network update period of $T_{net}$=10 ms. Thus, raw motion trajectories were first resampled to 100 Hz for model training. At each time instant, the predictor used historical motion states from the previous 10 $T_{net}$ and a target position sequence spanning from the previous 10 $T_{net}$ to 50 $T_{net}$ ahead to predict the velocity sequence over the next 10 $T_{net}$.

Two independent multilayer perceptron (MLP) based predictors were implemented.

One predictor was used to estimate the motion of the remote partner, whereas the other was used to estimate the motion of the shared virtual object. For the user-motion predictor, the input included user position history, user velocity history, box position history, and the target position sequence. For the shared-object motion predictor, the input further included box velocity history in addition to these variables. For each predictor, the model parameters were optimized by minimizing the mean squared error between the predicted and measured future velocity sequences:

$$\theta_q{}^* = \underset{\theta_q}{\operatorname{argmin}} \sum_t \left\| G_{\theta_q}\left(\mathcal{J}_q(t)\right) - \mathcal{V}_q(t) \right\|_2^2, \quad q \in \{u, b\} \tag{5}$$

where $G_{\theta_q}$ denotes the MIL predictor, $\mathcal{J}_q$ is the input feature vector at time $t$, $\mathcal{V}_q(t)$ is the measured future velocity sequence, and where $q = u$ denotes user-velocity prediction, and $q = b$ denotes box-velocity prediction.

**Online motion prediction**

During online interaction, each local terminal used ILMP to compensate for the delayed motion state received from remote terminals, as shown in Fig.3B. At each control cycle, the latest received remote position and velocity were used to construct the ILMP input, and the model predicted a short-horizon future velocity sequence. At $t_0$, the ILMP outputs ten predicted velocity samples: $\hat{v}(t_1)$, $\hat{v}(t_2), \dots$ $\hat{v}(t_m)$, where $t_j = t_0 + jT_{net}$, $m$ is the number of prediction steps output by the ILMP, which was set to 10 in this study.

Because the networked state update operated at 100 Hz, whereas the local robot controller operated at 1000 Hz, the predicted velocity sequence was interpolated to match the control-loop frequency: $\hat{v}_1$, $\hat{v}_2$, … $\hat{v}_{m \cdot n}$, where $\hat{v}_j = \hat{v}(t_0 + jT_{loc})$, $n = T_{net}/T_{loc}$, $(n \in Z^+)$. The latency-compensated remote position was reconstructed by integrating the interpolated predicted velocity from the latest received position: $\hat{x}_1$, $\hat{x}_2$, … $\hat{x}_{m \cdot n}$, where $\hat{x}_j = x(t_0) + \sum_{r=1}^{j} \hat{v}_r T_{loc}$, where $x(t_0)$ is the latest received position. The motion states used in Equations (1) and (3) were determined based on the measured network latency $t_d$. Specifically, the predicted state corresponding to $t_0 + t_d$ was first used to compensate for the delayed remote state. Because the local robot controller operated at a higher frequency than the networked state update, subsequent predicted states were then used for the following $n$ control steps until new networked information was received. The reconstructed position was then used to update the local replica of the remote user or shared object for synchronization and haptic rendering. For comparison, the NMP directly used the delayed received position, whereas the CVMP extrapolated the remote position using the latest received velocity.

**Participants and experimental protocol**

The study was approved by the Ethics Committee of Southern University of Science and Technology under approval number 20230095. All experimental procedures were conducted in accordance with the approved study protocol and relevant institutional ethical guidelines. Human participants in this study included 284 healthy individuals and 111 patients with neurological impairments. Written informed consent was obtained from all participants before the experiments. Detailed information on the experimental implementation, participant allocation, and demographic and clinical characteristics is provided in the Supplementary Materials.

### Statistical analysis

Statistical analyses were conducted using MATLAB (R2025b). All statistical methods and results are described in detail in the Supplementary Materials. In addition, a more detailed statistical report and the raw data are provided as accompanying CSV files.

### References and Notes

1. Y. L. Zhao, D. D. Zhang, P. Y. Gao, Y. Fu, Y. J. Ge, H. C. Chi, Z. X. Guo, H. H. Yu, J. F. Feng, L. Tan, W. Cheng, Y. R. Zhang, J. T. Yu, Associations of social isolation and loneliness with neurological disorders, psychiatric disorders, brain structures and behavioural phenotypes among UK Biobank participants. *Nat. Commun.* (**2026**)
2. R. J. Stolwyk, T. Mihaljcic, D. K. Wong, J. E. Chapman, J. M. Rogers, Poststroke cognitive impairment negatively impacts activity and participation outcomes: A systematic review and meta-analysis. *Stroke* **52**, 748-760 (2021).
3. K. D. Anderson, A. M. Bryden, B. Gran, S. W. Hinze, M. A. Richmond, Definitions of recovery and reintegration across the first year: A qualitative study of perspectives of persons with spinal cord injury and caregivers. *Spinal Cord* **62**, 156-163 (2024).
4. C. W. Cené, T. M. Beckie, M. Sims, S. F. Suglia, B. Aggarwal, N. Moise, M. C. Jiménez, B. Gaye, L. D. McCullough, Effects of objective and perceived social isolation on cardiovascular and brain health: A scientific statement from the American Heart Association. *J. Am. Heart Assoc.* **11**, e026493 (2022).
5. M. Kanbay, C. Tanriover, M. E. Demir, S. Copur, B. Afsar, A. Covic, D. H. K. Nakagawa, Social isolation and loneliness: Undervalued risk factors for disease states and mortality. *Eur. J. Clin. Invest.* **53**, e14032 (2023).
6. A. Dhand, D. Luke, J.-M. Lee, Social networks and risk of delayed hospital arrival after acute stroke. *Nat. Commun.* **10**, 1206 (2019).
7. E. Tang, N. Moran, M. Cadman, S. Hill, C. Sloan, E. Warburton, on behalf of the guideline committee, Stroke rehabilitation in adults: Summary of updated NICE guidance. *BMJ* **384**, q498 (2024).
8. M. Kritz, H. Riddell, D. Olsen, S. M. Harden, S. M. Burke, et al., Individual versus group-based interventions: a systematic review and meta-analysis of physical activity, functional, psychosocial and health outcomes. *Nat. Hum. Behav.* **10**, 1109-1121 (2026).
9. M. Gittler, A. M. Davis, Guidelines for adult stroke rehabilitation and recovery. *JAMA* **319**, 820-821 (2018).
10. K. O'Connell, A. A. Marsh, A. Seydell-Greenwald, Right hemisphere stroke is linked to reduced social connectedness in the UK Biobank cohort. *Sci. Rep.* **14,** 27293 (2024).
11. M. Goršič, I. Cikajlo, D. Novak, Competitive and cooperative arm rehabilitation games played by a patient and unimpaired person: effects on motivation and exercise intensity. *J. Neuroeng. Rehabil.* **14**, 23 (2017).

12. K. Baur, P. Wolf, V. Novak, D. Boering, S. Horner, C. Dahlen, J. Berger, R. Riener, V. Hemberg, Competitive versus cooperative forms of therapeutic gaming with subacute stroke patients. *IEEE Trans. Med. Robot. Bionics* **5**, 964-973 (2023).
13. Y. Noteboom, A. W. A. Montanus, F. van Nassau, G. Burchell, J. R. Anema, M. A. Huysmans, Barriers and facilitators of collaboration during the implementation of vocational rehabilitation interventions: A systematic review. *BMC Psychiatry* **24**, 759 (2024).
14. F. Pereira, S. Bermúdez i Badia, C. Jorge, M. S. Cameirão, The use of game modes to promote engagement and social involvement in multi-user serious games: A within-person randomized trial with stroke survivors. *J. NeuroEng. Rehabil.* **18**, 62 (2021).
15. J. Tosto-Mancuso, K. C. Tabacof, S. Herrera, D. G. Putrino, Gamified neurorehabilitation strategies for post-stroke motor recovery: challenges and advantages. *Curr. Neurol. Neurosci. Rep*. **22**, 183–195 (2022).
16. C. English, S. Hillier, E. Lynch, Circuit class therapy for improving mobility after stroke. *Stroke* **48**, e275–e276 (2017).
17. T. Tasiemski, P. K. Urbański, S. Jörgensen, et al., Effects of wheelchair skills training during peer-led Active Rehabilitation Camps for people with spinal cord injury in Poland: A cohort study. *Spinal Cord* **62**, 651-657 (2024).
18. M. S. Hossain, V. D. Novak, Combining gamification and haptic coupling in a two-dimensional tracking task performed by human dyads. *IEEE Trans. Haptics* **18**, 732-741 (2025).
19. K. B. Reed, M. A. Peshkin, Physical collaboration of human-human and human-robot teams. *IEEE Trans. Haptics* **1**, 108-120 (2008).
20. E. Noohi, M. Zefran, J. L. Patton, A model for human-human collaborative object manipulation and its application to human-robot interaction. *IEEE Trans. Rob.* **32**, 880-896 (2016).
21. C. E. Madan, A. Kucukyilmaz, T. M. Sezgin, C. Basdogan, Recognition of haptic interaction patterns in dyadic joint object manipulation. *IEEE Trans. Haptics* **8**, 54-66 (2015).
22. M. Mace, N. Kinany, P. Rinne, A. Rayner, P. Bentley, E. Burdet, Balancing the playing field: Collaborative gaming for physical training. *J. NeuroEng. Rehabil.* **14**, 116 (2017).
23. S. J. Kim, Y. Wen, D. Ludvig, E. B. Kucuktabak, M. R. Short, K. Lynch, L. Hargrove, E. J. Perreault, J. L. Pons, Effect of dyadic haptic collaboration on ankle motor learning and task performance. *IEEE Trans. Neural Syst. Rehabil. Eng.* **31**, 416-425 (2023).
24. E. L. Waters, R. J. Mendonca, P. Z. Cacchione, M. J. Johnson, TheraDyad: Feasibility of an affordable robot for multi-user stroke rehabilitation, paper presented at the 2024 *10th IEEE RAS/EMBS International Conference for Biomedical Robotics and Biomechatronics (BioRob),* Heidelberg, Germany, 1-4 September 2024.
25. K. Antonakoglou, X. Xu, E. Steinbach, T. Mahmoodi, M. Dohler, Toward haptic communications over the 5G tactile internet. *IEEE Commun. Surv. Tutorials* **20**, 3034-3059 (2018).
26. W. T. M. Lin, B. S. Lin, I. J. Lee, S. H. Lee, Development of a smartphone-based mHealth platform for telerehabilitation. *IEEE Trans. Neural Syst. Rehabil. Eng.* **30**, 2682-2691 (2022).

27. Y. Liu, S. Guo, Z. Yang, H. Hirata, T. Tamiya, A home-based tele-rehabilitation system with enhanced therapist-patient remote interaction: A feasibility study. *IEEE J. Biomed. Health Inform.* **26**, 4176-4186 (2022).
28. H. Liang, S. Liu, Y. Wang, J. Pan, Y. Zhang, X. Dong, Multi-user upper limb rehabilitation training system integrating social interaction. *Comput. Graph.* **111**, 103-110 (2023).
29. E. L. Waters, M. J. Johnson, Motor learning in robot-based haptic dyads: A Review. *IEEE Trans. Haptics* **17**, 510-527 (2024).
30. E. B. Küçüktabak, M. R. Short, Y. Wen, K. Lynch, L. J. Hargrove, J. L. Pons, J. A. G. Hale, J. C. G. Doerger, K. D. Katyal, E. J. Perreault, N. S. Makowski, C. J. Nycz, Therapist-exoskeleton-patient interaction for gait therapy. *Sci. Robot.* **11**, eadz9628 (2026).
31. A. Takagi, G. Ganesh, T. Yoshioka, M. Kawato, E. Burdet, Physically interacting individuals estimate the partner's goal to enhance their movements. *Nat. Hum. Behav.* **1**, 54 (2017).
32. G. Ganesh, A. Takagi, R. Osu, T. Yoshioka, M. Kawato, E. Burdet, Two is better than one: Physical interactions improve motor performance in humans. *Sci. Rep.* **4**, 3824 (2014).
33. R. P. R. D. van der Wel, G. Knoblich, N. Sebanz, Let the force be with us: Dyads exploit haptic coupling for coordination. *J. Exp. Psychol. Hum. Percept. Perform.* **37**, 1420-1431 (2011).
34. A. Noccaro, S. Buscaglione, J. Eden, X. Cheng, N. Di Stefano, E. Burdet, D. Formica, robot-mediated asymmetric connection between humans can improve performance without increasing effort. *IEEE Trans. Biomed. Eng.* **72**, 2675-2683 (2025).
35. A. Michałko, F. Di Tommaso, E. Peperoni, M. Leman, Robot-mediated haptic feedback outperforms vision in violin duo coordination. *Sci. Robot.* **11**, eaeb1901 (2026).
36. E. J. Avila Mireles, J. Zenzeri, V. Squeri, P. Morasso, D. De Santis, Skill learning and skill transfer mediated by cooperative haptic interaction. *IEEE Trans. Neural Syst. Rehabil. Eng.* **25**, 832-843 (2017).
37. I. Marcantoni, R. Assogna, L. Burattini, Ratio indexes based on spectral electroencephalographic brainwaves for assessment of mental involvement: A systematic review. *Sensors* **23**, 5968 (2023).
38. L. E. Ismail, W. Karwowski, Applications of EEG indices for the quantification of human cognitive performance: A systematic review and bibliometric analysis. *PLoS ONE* **15**, e0242857 (2020).
39. Y. J. Kim, Anthropomorphic low-inertia high-stiffness manipulator for high-speed safe interaction. *IEEE Trans. Robot.* **33**, 1358-1374 (2017).
40. J. Wang, X. Li, T.-H. Huang, S. Yu, Y. Li, T. Chen, A. Carriero, M. Oh-Park, H. Su, Comfort-centered design of a lightweight and backdrivable knee exoskeleton. *IEEE Robot. Autom. Lett.* **3**, 4265–4272 (2018).
41. D. A. Winter, *Biomechanics and motor control of human movement* (Wiley, Hoboken, NJ, ed. 4, 2009).

**Acknowledgments:** The authors thank the participants for their involvement in this study and acknowledge the technical assistance provided by the laboratory staff.

**Funding:**
National Natural Science Foundation of China grant 62273173
National Key R&D Program of China grant 2023YFF1205200, 2025YFF0523600
Guangdong Major Project of Basic Research grant 2025B0303000003

**Author contributions:**
Conceptualization: MMZ, CYS
Methodology: CYS, HDD, YDL, JL
Investigation: CLH, JG
Experimental implementation: CYS, MJD, HDD, JL, CLH, JG, YTL, YL, JJL, ZX, SNZ, XMZ
Data curation: HDD, JL,
Formal analysis: CYS, HDD
Visualization: CYS, YDL, HDD
Funding acquisition: MMZ
Project administration: CYS, MMZ
Supervision: MMZ, MJD, JSD, HHL, DRW, ZHL, ZYW
Writing – original draft: CYS, MJD, HDD
Writing – review & editing: CYS, MMZ, MJD, JSD, HHL, DRW, ZHL, ZYW

**Competing interests:** Authors declare that they have no competing interests.

**Data and materials availability:** All data are available in the main text or the supplementary materials.